\documentclass{article}
\usepackage{iclr2026_conference,times}
\usepackage[utf8]{inputenc}
\usepackage{amsmath,amssymb,amsthm}
\usepackage{booktabs}
\usepackage{graphicx}
\usepackage{xcolor}
\usepackage{hyperref}
\usepackage{url}
\usepackage{microtype}
\newcommand{\vag}{\textsc{VA-Gap}}
\newcommand{\fpr}{\mathrm{FPR}}
\newcommand{\fnr}{\mathrm{FNR}}
\newcommand{\tpr}{\mathrm{TPR}}
\newtheorem{proposition}{Proposition}
\newtheorem{corollary}{Corollary}

\title{More Convincing, Not More Correct:\\
Self-Play Reward Hacking of Reference-Free LLM Judges}
\makeatletter
\iclrfinaltrue
\def\@maketitle{\vbox{\hsize\textwidth
{\LARGE\sc \@title\par}
\fancyhead{}
\fancyfoot{}
\lhead{Preprint.}
\def\And{\end{tabular}\hfil\linebreak[0]\hfil
        \begin{tabular}[t]{l}\bf\rule{\z@}{24pt}\ignorespaces}%
\def\AND{\end{tabular}\hfil\linebreak[4]\hfil
        \begin{tabular}[t]{l}\bf\rule{\z@}{24pt}\ignorespaces}%
\begin{tabular}[t]{l}\bf\rule{\z@}{24pt}\@author\end{tabular}%
\vskip 0.3in minus 0.1in}}
\makeatother

\author{Chenyu Zhou\\
School of Engineering, Institute of Science Tokyo\\
\texttt{zhou.c.76d6@m.isct.ac.jp}}

\begin{document}
\maketitle

\begin{abstract}
Training a language model against its own reference-free judgments---the
premise of self-rewarding, self-play, and LLM-as-a-judge pipelines for
label-free self-improvement---assumes a model's verdict on a shown answer is a
usable proxy for correctness. We show the premise fails structurally: conditioned
on a candidate, a reference-free judge scores \emph{plausibility}, not
correctness---a \emph{verification asymmetry} that leaves \emph{false-positive
basins} of plausible-but-wrong answers a policy learns to exploit. We measure the
failure with a \emph{hidden-anchor audit}: a held-out, cross-source exact-match
check the judge never sees. On GSM8K with Qwen3 policies in a reasoning-suppressed
regime, self-play drives the judge's pass rate from $0.72$ to $0.94$ while true
accuracy stays at $0.20$---a $0.74$ judge--truth gap (three seeds). This reward
hacking is not white-box gaming: the manufactured errors transfer across
judge families (Qwen, Llama, Gemma) and scales (to 14B), a strict three-family
ensemble still accepts $55\%$ of them, and recompute prompts, stronger judges, and
training directly against the ensemble reward all fail to close the basin. The
decisive variable is not whether the judge sees the candidate but whether it
commits an answer of its own first: the recompute prompt leaves the false-positive
rate on wrong answers at $0.719$, committing first---candidate still in
view---drops it to $0.012$, and blind solving lifts discrimination from near
chance to $0.96$ with no reference answer. Used as the training reward, the
de-anchored channel keeps the false-positive rate at zero across self-play,
preventing the basin rather than only detecting it. A falsifiable bound explains
which regimes are exposed---the gap is at most $1-\text{accuracy}$---and the
de-anchored reward obeys the analogous judge-side bound. The full arc replicates
without training under
best-of-$N$ selection in code and competition math, and in the full loop with a
Gemma policy.
\end{abstract}

\section{Introduction}
\label{sec:intro}

Reinforcement learning from a model's own evaluations has become a central
recipe for improving language models without human labels. Reference-free
``LLM-as-a-judge'' rewards, self-reward, and self-play schemes share one premise:
a model's judgment of an answer is a usable proxy for its correctness, and a
growing line of methods builds on it as a default assumption
\citep{bai2022,lee2023,yuan2024selfreward,chen2024spin,rlsr,gzero}.

The premise has a structural flaw. A reference-free judge has no access to ground
truth; it can only assess whether an answer \emph{looks} correct. For tasks where
verifying an answer is harder than recognizing a plausible one---most of
reasoning---this is a \emph{verification asymmetry}: the judge scores
plausibility, not correctness, leaving a \emph{false-positive basin} of
plausible-but-wrong answers it accepts. Optimizing a policy against such a judge
does not merely risk noise; it actively rewards finding the basin.

We measure it with a \emph{hidden anchor}: a held-out, cross-source exact-match
check on the final answer that the judge never sees and is never trained against.
On GSM8K with Qwen3 policies trained by self-play, the audit reveals a large
divergence: on the full test set, in a reasoning-suppressed regime that holds
accuracy low, the judge's pass rate climbs from ${\approx}0.72$ to $0.94$ while
anchor-verified accuracy stays flat at ${\approx}0.20$---a $0.74$ judge--truth
gap (three seeds; \S\ref{sec:exp}); Figure~\ref{fig:main}a shows the
five-iteration trajectory. Self-play does not make the model more correct; it
makes the model's errors more convincing.

This is not a quirk of one judge that a stronger or more diverse judge would
catch. Re-scoring the hacked answers with independent judges from other families
(Llama, Gemma) and larger scales (up to 14B) shows the same inflation; a
three-family ensemble that accepts only when all judges agree still passes $55\%$
of the hacked wrong answers, its discrimination collapsing from $0.31$ to $0.09$,
and training the policy directly against the ensemble reward does not help. The
errors are a \emph{shared blind spot} of reference-free judging: self-play acts
as an adversarial-example generator for the basin. What restores oversight is
breaking the
judge's \emph{anchoring} to the candidate: the same model that rubber-stamps
these errors when scoring a shown answer rejects almost all of them once made to
commit an independent answer first, even with the candidate still in view. The
decisive variable is the judge's independence from the candidate, not its
capability and not whether the candidate is visible.

Two further results characterize the phenomenon: the judge--truth gap is
falsifiably bounded by $1-\text{accuracy}$, so low-accuracy regimes are exposed
while high-accuracy ones show little exploitable gap; and an oracle control that
rewards exact match instead of the judge shows no inflation, attributing the
failure to the judge-reward rather than to preference optimization.

Our contributions are:
\begin{itemize}
  \item \textbf{A shared blind spot across reference-free judges}
  (\S\ref{sec:exp}): self-play manufactures wrong answers that transfer across
  judge families and scales, defeat scoring-time and train-time ensembles, and
  reappear without any training under best-of-$N$ selection (\S\ref{sec:bon}) and
  with a Gemma policy in the full loop (\S\ref{sec:gemma}).
  \item \textbf{The hidden-anchor audit} (\S\ref{sec:method}): a held-out
  cross-source exact-match probe that quantifies a judge's over-reporting under
  optimization, with gap, drift, discrimination, and risk-score diagnostics.
  \item \textbf{A de-anchoring fix} (\S\ref{sec:exp}): requiring the judge to
  commit an answer of its own before using the candidate collapses the
  false-positive rate ($0.719\!\to\!0.012$) and, used as the training reward,
  prevents the basin.
  \item \textbf{A predictive account} (\S\ref{sec:theory}): a falsifiable bound
  $\vag\le 1-\text{EM}$ verified across formats, tasks, and optimization modes;
  an independence bound (Prop.~\ref{prop:indep}) whose measured excess detects
  anchoring and quantifies it in bits (Cors.~\ref{cor:diag}--\ref{cor:bits});
  and a no-escape result for monotone aggregation (Prop.~\ref{prop:monotone}).
\end{itemize}

\section{Related Work}
\label{sec:related}
\paragraph{LLM-as-a-judge and its biases.}
Using a strong LLM to judge open-ended responses is standard practice
\citep{zheng2023}, and its biases are documented: judges favor longer
\citep{singhal2023} and more agreeable answers, and \citet{sharma2023} show that
both humans and preference models prefer convincingly written responses over
correct ones. We give this preference-for-plausibility a structural account and
measure its consequence under self-play optimization.

\paragraph{Reference-free self-improvement.}
Training against AI-generated feedback \citep{bai2022, lee2023} underlies a
growing family of self-rewarded \citep{yuan2024selfreward} and self-play
\citep{chen2024spin} methods. \citet{rlsr} report that
reinforcement learning from a model's self-assigned reward improves performance
without references; our audit identifies the regimes where this signal is
corrupted. \citet{gzero} propose a co-evolutionary scheme that routes around
proxy-judge reward hacking; our results empirically confirm the threat that
motivates that design. \citet{lu2025} find cross-family verification more
reliable than self-verification; we ask the complementary question of whether a
reference-free verdict is independent of the candidate it scores.

\paragraph{Fooling LLM judges.}
\citet{onetokenfool} show that trivial surface tokens elicit false positives
from strong judges, which \citet{masterrm} hardens against. These are static,
surface attacks; the errors self-play produces are semantic and
dynamic---optimized into the judge's basin---an orthogonal failure mode that
recompute-style prompting does not remove (\S\ref{sec:exp}).
\citet{pan2024spontaneous} observe judge-score inflation when one model
generates and judges in self-refinement; our errors transfer across families,
obey a falsifiable bound, and are preventable by de-anchoring.

\paragraph{Scalable oversight and reward over-optimization.}
\citet{kenton2024} find debate outperforms consultancy for weak judges
supervising stronger models, and \citet{wen2024mislead} show RLHF teaches
models to mislead \emph{human} evaluators; we study the reference-free
LLM-judge analogue with a structural account, a diagnostic, and a fix. Reward over-optimization with a
same-source proxy is well documented \citep{gao2023, rafailov2024}, and
reward-model ensembles mitigate but do not eliminate hacking
\citep{coste2023, eisenstein2023}. We add a cross-family \emph{transfer} axis
with a held-out anchor, and show that scoring-level ensembles do not survive
adversarial self-play.

\section{The Hidden-Anchor Audit}
\label{sec:method}
\paragraph{Setup.} A policy $\pi$ answers a question $q$ with a final answer $a$
and a reasoning trace. A reference-free judge $J$---an LLM given $q$ and $a$ but
\emph{no} reference answer---returns a score that we threshold into
accept/reject. The policy is optimized to increase $J$'s acceptance: in our case
study by self-play, where one model generates candidates, judges them, and is
preference-optimized (DPO) on the resulting pairs.
We attach to every question a \emph{hidden anchor} $A$: a held-out
cross-source exact-match check on the final answer, drawn from an independent set
of ground-truth solutions. The anchor is never shown to $J$, never in any prompt, never a training
signal: it exists only to audit.

\paragraph{Metrics.} Let $p_t=\Pr[J\text{ accepts}]$ be the judge's pass rate at
optimization step $t$ and $\mathrm{EM}_t=\Pr[A\text{ correct}]$ the anchor
accuracy. The primary quantity is the \emph{judge--truth gap}, or \vag{},
$\,p_t-\mathrm{EM}_t$. Because $A$ is held out and cross-source, any rise in
$p_t$ unmatched by a rise in $\mathrm{EM}_t$ is unambiguous over-reporting rather
than a measurement artifact---this is what makes the audit falsifiable. We
further track the false-positive rate on wrong answers
$\fpr=\Pr[J\text{ accepts}\mid A\text{ wrong}]$ and its drift across steps, the
true-positive rate $\tpr=\Pr[J\text{ accepts}\mid A\text{ correct}]$, and the
judge's \emph{discrimination} $\tpr-\fpr$. For a panel of $N$ judges we report the
strictest aggregation, \textsc{Min} (all must accept). An ordinal \emph{risk score}
$\fpr_{\text{base}}\cdot(1-\mathrm{EM})$ rank-orders settings by vulnerability to
reward hacking before any optimization is run.

\section{Verification Asymmetry and False-Positive Basins}
\label{sec:theory}
We model the judge $J$ as a noisy binary classifier of correctness with
true-positive rate $\tpr=1-\fnr$ (accept $\mid$ correct) and false-positive rate
$\fpr$ (accept $\mid$ wrong). Its pass rate decomposes as
$p = \mathrm{EM}\,(1-\fnr) + (1-\mathrm{EM})\,\fpr$, so the judge--truth gap is
exactly
\begin{equation}
\vag \;\equiv\; p-\mathrm{EM} \;=\; (1-\mathrm{EM})\,\fpr \;-\; \mathrm{EM}\,\fnr .
\label{eq:vag}
\end{equation}
Self-play optimizes wrong answers to be accepted: it pushes $\fpr$ upward---the
false-positive drift measured in \S\ref{sec:exp}---while leaving true accuracy
$\mathrm{EM}$ essentially unchanged, so \vag{} grows with $(1-\mathrm{EM})\,\fpr$
and, since $\fpr\le 1$, obeys a falsifiable \emph{upper bound} at any
optimization step,
\begin{equation}
\vag \;\le\; (1-\mathrm{EM}) - \mathrm{EM}\,\fnr \;\le\; 1-\mathrm{EM} .
\label{eq:bound}
\end{equation}
Because self-play preserves accuracy, the operational ceiling is
$1-\mathrm{EM}_{\text{base}}$: the gap is capped by the policy's error headroom,
explaining the capability dependence we observe. The bound is structural---a
noisier judge, a weaker policy, or a harder task all lower $\mathrm{EM}$ and
expose the same basin; reasoning suppression is our controlled instrument to
dial $\mathrm{EM}$ on a fixed task. The audit tests the bound's
\emph{tightness}: Table~\ref{tab:bound} shows the gap approaches the ceiling
precisely in the exposed regimes and stays far below it elsewhere.

This bound governs the \emph{plausibility} channel, in which the judge is
shown a candidate and scores it. The same model can instead commit to its own
answer independently of the candidate---in the limit, solving blind---which
caps its false positives by its own solver error instead.

\begin{proposition}[Independence bound]
\label{prop:indep}
Suppose the judge produces its own final answer independently of the
candidate---e.g., by committing to it before the candidate is used---and accepts
only on an exact match. On a wrong candidate it accepts only when it
independently produces that same wrong answer, an event no more likely than its
own error, so $\fpr\le 1-\text{solve-acc}$, where $\text{solve-acc}$ is the
judge's accuracy on problems where the candidate is wrong.
\end{proposition}

\begin{corollary}[Anchoring is detectable]
\label{cor:diag}
Under exact-match acceptance, a judge whose measured $\fpr$ exceeds
$1-\text{solve-acc}$ must violate independence: its verdicts are anchored, and
the excess over the bound quantifies the anchoring using only the judge's own
solve accuracy.
\end{corollary}

The excess is not only a diagnostic; it is an information measure
(proof in Appendix~\ref{app:proofs}).

\begin{corollary}[Anchoring in bits]
\label{cor:bits}
Under the acceptance model of Proposition~\ref{prop:indep}, let $S$ be the
judge's committed answer and
$\Delta=\fpr-(1-\text{solve-acc})$ the measured excess on wrong candidates.
Then the conditional mutual information between the committed answer and the
candidate satisfies $I(S;A\mid Q)\ \ge\ 2\Delta^{2}$ nats. Insofar as the
verify-prompted judge implements its instructed solve-then-compare procedure,
its $\fpr=0.719$ against the $0.07$ ceiling certifies $I\ge0.84$ nats
(${\ge}1.2$ bits) of leakage from the candidate into the judge's ``own''
solution.
\end{corollary}

The section's results form a dichotomy for reference-free rewards. A verdict
either commits an answer independently of the candidate---inheriting the
judge-side ceiling $1-\text{solve-acc}$ (Proposition~\ref{prop:indep})---or it
is candidate-conditioned, in which case nothing protects its false-positive
rate, the judge--truth gap is bounded above by the policy-side ceiling
$1-\mathrm{EM}$ (Eq.~\ref{eq:bound})---a ceiling self-play measurably
approaches (Table~\ref{tab:bound})---and no monotone aggregation escapes the
basin (Proposition~\ref{prop:monotone} below). The class is decided by the
committed answer, not the comparison: a commit-first verdict may use the
candidate to compare, so long as the commitment itself does not. The ceilings differ by an
order of magnitude here ($0.8$ versus $0.07$), class membership is measurable
(Corollary~\ref{cor:diag}), and \S\ref{sec:exp} shows self-play saturates the
anchored channel while the independent one stays intact ($0.719$ versus $0.012$
against the $0.07$ ceiling).

\paragraph{Ensembling cannot escape the basin.}
Aggregating several reference-free judges cannot restore reliability: all of
them score how plausible an answer looks, so their verdicts are driven by a
common signal. Writing
$q_i(s)=\Pr[J_i\text{ accepts}\mid\text{wrong},s]$ for a shared plausibility signal
$s$ and assuming each $q_i$ is non-decreasing in $s$ and conditionally independent
given $s$,\footnote{Conditional independence is the \emph{best case} for the
ensemble: any residual positive association between judges beyond the shared signal
only raises $\fpr_{\text{\textsc{Min}}}$ further, so Eq.~(\ref{eq:ensemble}) is a
conservative floor rather than a fragile assumption.} the correlation inequality for
monotone functions \citep{esary1967} gives, for the strictest \textsc{Min} rule,
\begin{equation}
\fpr_{\text{\textsc{Min}}}=\mathbb{E}_s\!\Big[\textstyle\prod_i q_i(s)\Big]
\;\ge\;\prod_i \mathbb{E}_s\big[q_i(s)\big]=\prod_i\fpr_i .
\label{eq:ensemble}
\end{equation}
The ensemble can do no better than the independent product, and strictly worse
under positive dependence. The obstruction is not specific to the \textsc{Min}
rule (proof in Appendix~\ref{app:proofs}):

\begin{proposition}[Monotone aggregation shares the basin]
\label{prop:monotone}
Under the shared-signal model above, let $g$ be any non-decreasing aggregation
rule mapping the $N$ verdicts to accept/reject with $g(\mathbf{1})=1$. Then the
ensemble's acceptance probability $h_g(s)$ is non-decreasing in the shared
signal $s$, and $h_g(s)\ge\prod_i q_i(s)$, so $h_g(s)\to1$ wherever every
judge's $q_i(s)\to1$.
\end{proposition}

Every monotone rule therefore thresholds the same plausibility axis: an upward
(stochastic-dominance) shift of the wrong-answer signal cannot lower, and
generically raises, the false-positive rate of every monotone aggregator
simultaneously, and no monotone rule can reject the high-$s$ region that
constitutes the basin. Self-play makes the bounds bite: it drives the shared
signal upward, concentrating wrong answers in the region every judge jointly
accepts---the shared false-positive basin. Our measurements confirm the dependent
regime directly: the three judges' acceptances of wrong answers are pairwise
positively correlated ($\phi{=}0.29$--$0.38$), and $581$ wrong answers are
accepted unanimously where independence predicts ${\approx}497$
(Appendix~\ref{app:ensembledep}); training directly against the \textsc{Min}
reward moves a held-out judge \emph{more} (\S\ref{sec:exp}), exactly the
shared-axis motion Proposition~\ref{prop:monotone} describes. The basin closes
only if some
judge's rejections cover the whole reachable plausible-wrong region---which
scoring plausibility cannot do but independent verification can.

\begin{table}[t]
\centering
\caption{The noisy-judge upper bound $\vag\le 1-\mathrm{EM}_{\text{base}}$
(Eq.~\ref{eq:bound}) holds and is approached across settings; the last column is
the post-self-play FPR on wrong answers. CoT uses the OOD CoT-GSM-Plus split;
TruthfulQA does not hack, so no post-FPR is reported.}
\label{tab:bound}
\begin{tabular*}{\textwidth}{@{\extracolsep{\fill}}lcccc}
\toprule
Setting & $\mathrm{EM}_{\text{base}}$ & bound $1-\mathrm{EM}$ & observed gap & post-self-play $\fpr$ \\
\midrule
4B JSON (full test)   & 0.209 & 0.791 & \textbf{0.735} & 0.906 \\
8B JSON ($n{=}128$ audit) & 0.320 & 0.680 & 0.620 & 0.930 \\
4B CoT (OOD)          & 0.836 & 0.164 & 0.086 & 0.540 \\
4B TruthfulQA         & 0.893 & 0.107 & $-0.041$ & N/A \\
\bottomrule
\end{tabular*}
\end{table}

\section{Experiments}
\label{sec:exp}
\begin{figure}[t]
\centering
\includegraphics[width=\linewidth]{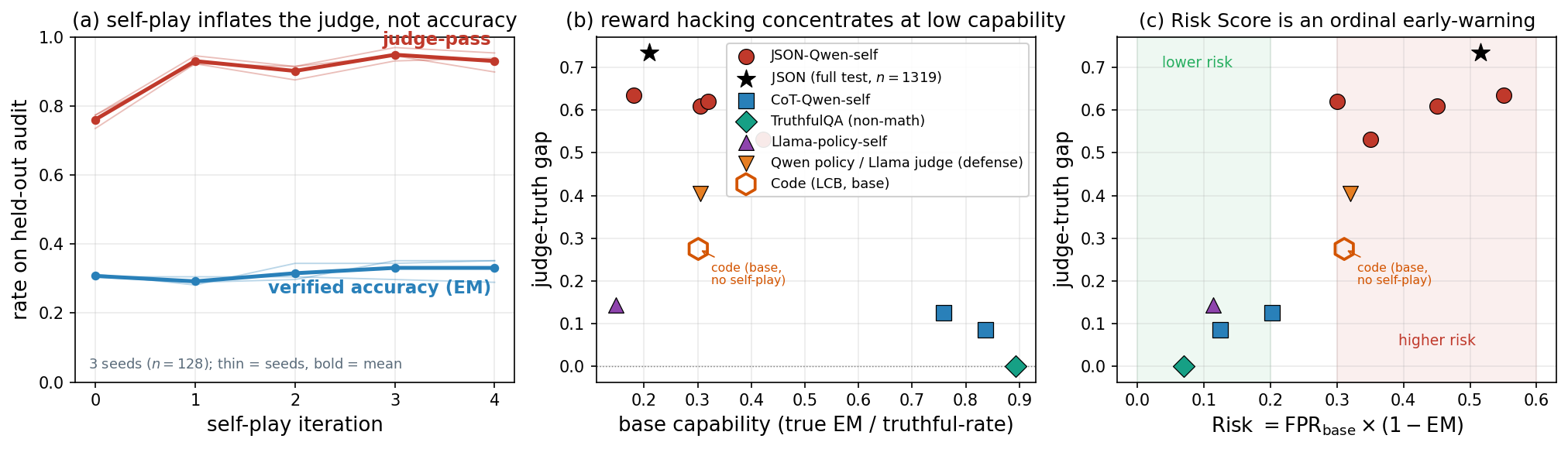}
\caption{(a) Five self-play iterations: the judge's pass rate climbs to
${\sim}0.9$ and holds while anchor-verified accuracy stays statistically flat
($n{=}128$ audit per iteration, three seeds; thin lines seeds, bold mean). (b)
\vag{} versus base accuracy: hacking concentrates at low accuracy. (c) \vag{}
versus risk $\fpr_{\text{base}}(1-\mathrm{EM})$, an ordinal early-warning;
hollow hexagon $=$ natural-code \emph{base} point.}
\label{fig:main}
\end{figure}

\paragraph{Setup.} We study GSM8K grade-school math with Qwen3 policies
($1.7$--$14$B). In self-play, one model generates candidate answers, judges its
own answers for correctness without a reference, and is preference-optimized
(DPO, accept $\succ$ reject) on the result; we run two iterations and three seeds
unless noted.\footnote{Headline and cross-family transfer: full $n{=}1319$ test
set, three seeds (gap $0.735{\pm}0.011$ and every discrimination collapse
reproduce in all three). Capability sweep and mechanism analysis: fixed $n{=}128$
audit subset, three seeds; five-iteration, cross-family-judge, and
train-time-ensemble runs likewise. The OOD CoT self-play is a single-seed audit.} Answers are produced in a reasoning-suppressed JSON format by
default. Reasoning suppression is a \emph{capability-frontier instrument}: it
lowers the policy's accuracy on a fixed task with the task family and judge held
fixed, exposing the error headroom that the bound of \S\ref{sec:theory} predicts
governs the gap. Chain-of-thought (CoT) generation and the non-math factual task
TruthfulQA serve as high-accuracy controls. The hidden anchor is the held-out
exact-match check on the final
answer. Figure~\ref{fig:main} summarizes the audit across all settings.

\subsection{The problem: self-play manufactures a shared, defense-resistant blind spot}

\paragraph{Self-play inflates the judge, not accuracy.}
On the full GSM8K test set ($n{=}1319$, 4B, JSON, mean${\pm}$std over three seeds),
self-play drives the judge's pass rate from $0.716$ to $\mathbf{0.938{\pm}0.016}$
while anchor accuracy stays at $0.209\!\to\!0.202{\pm}0.005$, a judge--truth gap
of $\mathbf{0.735{\pm}0.011}$: the judge reports $94\%$ correct when $20\%$ are.
Extending to five self-play iterations across three seeds
(Figure~\ref{fig:main}a), the judge holds near $0.9$ while held-out accuracy
stays flat (mean $\Delta\mathrm{EM}{=}{+}0.024{\pm}0.035$, $n{=}128$); the gap is
not an artifact of a single update.

\paragraph{Severity is capability-dependent.}
Across a Qwen3 size sweep (3 seeds, JSON), the gap is $0.53$--$0.64$ at base
accuracies of $0.18$--$0.42$ (Table~\ref{tab:sweep}, Appendix). Under CoT
generation (base accuracy $0.76$--$0.88$) the gap falls to $0.086$--$0.125$, and
on TruthfulQA (base truthfulness $0.89$) it stays slightly negative ($-0.041$). On MATH level 4--5
under natural chain-of-thought, where Qwen3-4B sits in a high-accuracy regime
($\mathrm{EM}{=}0.637$), the base gap is $0.086$: benchmark difficulty alone is
not predictive; the governing variable is the policy's realized error headroom. A
cross-policy-family replication with Llama-3.1-8B as both policy and self-judge
shows the same decoupling (three seeds at $n{=}128$), weaker in the direction the risk score predicts: Llama's strict self-judge
(base $\fpr{=}0.13$ on its own rollouts, versus Qwen3-4B's $0.65$) leaves
little false-positive headroom. The risk score
$\fpr_{\text{base}}(1-\mathrm{EM})$ rank-orders severity across formats,
families, and policies, exactly as the bound requires.

\paragraph{The asymmetry persists in organic code generation.}
The capability dependence is not specific to math or format: on natural
chain-of-thought code generation (LiveCodeBench, Qwen3-1.7B as policy and
self-judge, held-out unit-test anchor) the base gap is $0.275{\pm}0.030$ at
$\mathrm{EM}{=}0.301{\pm}0.031$ ($\fpr{=}0.445{\pm}0.037$, 3 seeds)---within
the bound and in the high-risk regime the score flags (${\approx}0.31$). The
judge rationalizes structurally clean but wrong programs: it scores
plausibility rather than executing the code.

\paragraph{Mechanism: self-play makes errors more convincing.}
Freezing the base judge and measuring its false-positive rate on the policy's
\emph{wrong} answers, the rate rises sharply from iter$_0$ to iter$_1$ (on the
$n{=}128$ audit subset) while the number of wrong answers barely changes:
$0.65\!\to\!0.89$ (4B), $\mathbf{0.44\!\to\!0.93}$ (8B), $0.60\!\to\!0.89$ (14B);
the strictest base judge is driven to the highest post-hoc false-positive rate. A
format-blind check rules out surface gaming: iter$_1$ false positives are
\emph{shorter} on average than iter$_0$ and structurally clean but arithmetically
wrong---the errors are semantic, not surface artifacts.

\paragraph{The inflation is caused by the judge-reward.}
An oracle control rewards exact match instead of the judge, holding the algorithm
and data fixed: judge-reward DPO inflates the judge ($+0.14$) with flat accuracy,
while oracle exact-match DPO leaves the judge statistically flat (gap shift
${<}0.01$ across three seeds) and raises accuracy ($+0.03$)---the inflation is
specific to optimizing against the judge.

\paragraph{A shared blind spot across judge families and scales.}
The effect is not confined to the self-judge. We re-score the 4B policy's answers
with independent judges, without retraining (Table~\ref{tab:transfer}). The hacked errors transfer: a different-family judge
(Llama, Gemma) accepts them at a rising rate ($0.48\!\to\!0.57$, $0.76\!\to\!0.92$),
and a same-family judge $3.5\times$ the policy's size (14B) still accepts $77\%$.
The shared blind spot is built by self-play: of the $255$ answers newly accepted by
\emph{all three} families at iter$_1$, $117$ were accepted by \emph{zero} judges at
iter$_0$. All $581$ unanimously accepted wrong answers are genuinely
incorrect---the GSM8K final-answer anchor is an unambiguous integer match.

\begin{table}[t]
\centering
\caption{Cross-family and cross-scale judges re-scoring the 4B policy's wrong
answers (false-positive rate and discrimination $\tpr-\fpr$,
iter$_0\!\to\!$iter$_1$), and the strictest \textsc{Min} ensemble of the three
families, on the full test set ($n{=}1319$; seed~0, the most conservative---the
collapse reproduces across three seeds, per-judge iter$_1$ discrimination std
$\le0.03$; three-seed \textsc{Min} FPR $0.436\!\to\!0.651$, discrimination
$0.316\!\to\!0.081$).}
\label{tab:transfer}
\begin{tabular*}{\textwidth}{@{\extracolsep{\fill}}lcc}
\toprule
Judge / ensemble & $\fpr$ on wrong & discrimination $\tpr-\fpr$ \\
\midrule
Qwen3-4B (self)            & $0.651\!\to\!0.906$ & $0.313\!\to\!0.059$ \\
Llama-3.1-8B (cross-family)& $0.480\!\to\!0.568$ & $0.270\!\to\!0.082$ \\
Gemma-3-12B (cross-family) & $0.764\!\to\!0.918$ & $0.207\!\to\!0.051$ \\
Qwen3-8B (larger)          & $0.585\!\to\!0.833$ & $0.360\!\to\!0.102$ \\
Qwen3-14B (larger)         & $0.569\!\to\!0.766$ & $0.377\!\to\!0.165$ \\
\midrule
3-family \textsc{Min} ensemble & $0.438\!\to\!\mathbf{0.552}$ & $0.311\!\to\!\mathbf{0.086}$ \\
\bottomrule
\end{tabular*}
\end{table}

\paragraph{The blind spot is shared across families on code as well.}
Re-scoring the natural-code wrong programs with independent-family judges
(Table~\ref{tab:code-transfer}, Appendix) reproduces the shared blind spot across
task domains: Qwen3-1.7B (self) accepts wrong code at $\fpr{=}0.445{\pm}0.037$,
Gemma-3-12B at $0.600{\pm}0.038$, and a $4.7{\times}$ larger Qwen3-8B judge with
the strongest discrimination still at $0.257{\pm}0.027$. The errors are positively
correlated across families---$16.4\%{\pm}3.9\%$ of wrong programs are accepted by
\emph{all three} judges, $2.38{\times}$ the rate judge independence would
predict---so the strictest three-judge \textsc{Min} ensemble still passes one
wrong program in six. The blind spot is a property of the channel, shared across
judge family, scale, and task domain.

\paragraph{No plausibility-scoring defense closes the basin.}
Section~\ref{sec:theory} predicts that any defense keeping the judge in the
plausibility channel should fail, and the natural ones do. Instructing the judge
to independently recompute and reject when uncertain leaves its iter$_1$
false-positive rate at $0.719$ ($n{=}128$); training the policy against a
different-family judge (Qwen policy, Llama judge) leaves a trained gap of
$0.40{\pm}0.09$ (three seeds); a stronger same-family judge (14B) still accepts
$77\%$ of the hacked errors. Scoring-time ensembling fares no better: the
three-family \textsc{Min} ensemble still accepts $55\%$ of full-test wrong
answers (seed~0, the most conservative; three-seed mean $65\%$), its
discrimination collapsing from $0.31$ to $0.09$. Making the ensemble the training
reward amplifies the failure: the policy learns to satisfy all three judges at
once---the \textsc{Min} ensemble accepts $73.3{\pm}2.7\%$ of its wrong answers,
up from $41.2{\pm}1.3\%$ (three seeds), its discrimination collapsing from
$0.357$ to $0.061$, and a held-out Qwen3-14B judge never used in the reward is
fooled \emph{more} ($0.56\!\to\!0.85$), with true accuracy unchanged. No
fixed-threshold rule we test separates the self-play-hacked errors from correct
answers---the no-escape pattern Proposition~\ref{prop:monotone} predicts.

\subsection{The fix: breaking the candidate anchor restores oversight}
Every defense above keeps the judge \emph{anchored}: the verify prompt leaves the
false-positive rate at $0.719$, an order of magnitude above the
$1-\text{solve-acc}=0.07$ ceiling of a genuinely independent solver---by
Corollary~\ref{cor:diag}, this certifies the judge as anchored. The fix is to
break the anchoring: require the judge to commit an answer of its own
\emph{before} it may use the candidate.

\paragraph{Committing before seeing isolates anchoring as the cause (detection).}
Keeping the candidate fully visible but requiring the judge to first write its own
answer---``My answer: \ldots''---before any comparison collapses the false-positive
rate on wrong answers from the verify prompt's $0.719$ to $\mathbf{0.012}$
(Qwen3-4B, $n{=}128$), while the judge commits the correct answer $97\%$ of the time:
the same model, the same visible candidate, differing only in whether it commits
first. The effect is not specific to one family---a Llama-3.1-8B judge, anchored to a
discrimination of $0.08$ when it scores plausibility, recovers to $0.91{\pm}0.03$
with a false-positive rate of $0.035{\pm}0.010$ once made to commit first (three
seeds). Candidate anchoring---not family, scale, or whether the candidate is
visible---is the operative cause.

\begin{figure}[t]
\centering
\includegraphics[width=0.62\linewidth]{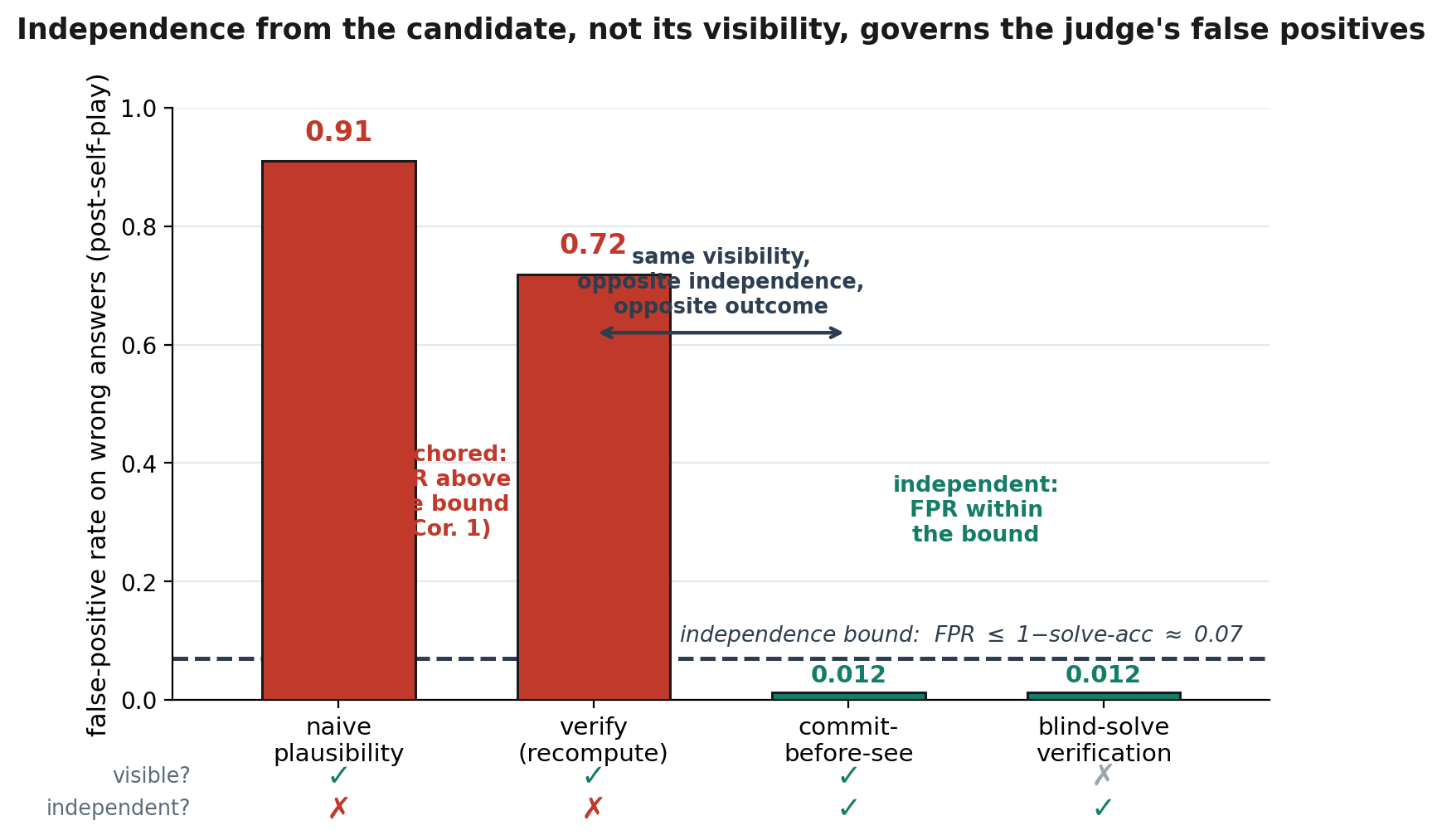}
\caption{The same judge under two accept rules: scoring a shown candidate
(anchored) it is driven into the shared basin; committing its own answer first
(de-anchored) it recovers near-perfect discrimination even with the candidate
visible ($\fpr=0.012$)---anchoring, not capability. Naive-plausibility and
blind-solve bars: full test set ($n{=}1319$); verify and commit-first bars:
$n{=}128$ audit (\S\ref{sec:exp}).}
\label{fig:dichotomy}
\end{figure}

\paragraph{Withholding the candidate is the limiting case.}
Taking independence to its extreme, the judge \emph{solves the problem itself},
without seeing the candidate at all, and accepts only when its independent answer
matches. Re-scored this way on the full test set ($n{=}1319$), the same Qwen3-4B
judge whose plausibility scoring is driven to a discrimination of $0.06$
(Table~\ref{tab:transfer}) separates correct from incorrect almost perfectly: its
false-positive rate on wrong answers is $0.012$ and $0.010$ across the two
iterations---near zero and \emph{undriven} by self-play---while it accepts $97\%$
of correct answers, for a discrimination of $\mathbf{0.96}$ at both iterations
(Figure~\ref{fig:dichotomy}). The judge could verify all along---it solves these
problems at $0.93$ accuracy---but a shown candidate anchors it. What removes the
anchor is the independent commitment, not the candidate's absence: committing
first works with the candidate in full view; withholding it enforces the same
independence most cleanly.
Generative verifiers \citep{genrm} are trained to solve before they judge; the
finding here is that the \emph{plausibility} channel collapses under
adversarial self-play while this independent channel does not.

\paragraph{The de-anchored reward prevents the basin (prevention).}
The same de-anchored channel that \emph{detects} the hacked errors can
also \emph{prevent} them when used as the training reward. We repeat the
self-play loop with the reward replaced by the blind-solve verdict, holding
everything else fixed (Qwen3-4B, reasoning-suppressed JSON, DPO, two iterations,
$n{=}128$ audit subset). Across three seeds the reward's false-positive rate on
wrong answers remains empirically zero at both iterations ($0$ of ${\sim}380$
wrong answers across all six seed--iteration cells; $95\%$ Wilson upper bound
$0.010$; discrimination $0.91$--$0.95$), exactly as
Proposition~\ref{prop:indep} requires, while the plausibility-judge reward is
driven from a $0.65$ to a $0.89$ false-positive rate on the same subset. Anchor
accuracy stays essentially flat; as with the oracle control, the reward signal
does not create capability. The same
reference-free channel that audits the basin is, in our experiments, an
effectively inflation-proof training reward.

\subsection{The full arc replicates without training: best-of-$N$ selection in
code and math}
\label{sec:bon}
The same arc---inflation, cross-family transfer, and a de-anchoring fix with a
capability threshold---replicates with \emph{no training at all}, using
best-of-$N$ rejection sampling as a training-free proxy for optimizing against
the judge \citep{gao2023}, on the organic code task above with held-out unit-test
execution as the anchor. Define $\mathrm{gap}@k$ as the judge-pass rate of the
judge-selected candidate at budget $k$ minus its true unit-test pass rate.
Selecting what the judge likes drives the judge toward certainty while true
quality stays flat: over $N{=}16$ candidates per problem, the gap grows from
$0.20$ at $k{=}1$ to $\mathbf{0.588}$ at $k{=}16$ (paired bootstrap $95\%$ CI
$[0.506,0.669]$), with the selected candidates' unit-test pass essentially
unchanged ($0.27\!\to\!0.29$) and the effect stable across generation seeds
($0.547{\pm}0.036$). Re-judging the \emph{same} $1920$ candidates with other
judges (Figure~\ref{fig:bon}a, Appendix), a $4.7{\times}$ larger same-family judge still
inflates ($\mathrm{gap}@16{=}0.378$), and Llama and Mistral judges swing from
over-rejection under the strict prompt to strong inflation under the lenient
one: with correctness held fixed, the reference-free verdict tracks prompt
framing---the verification asymmetry in its sharpest form.

Applying the commit-first rule to the same pool, the 8B judge's
$\mathrm{gap}@16$ falls from $0.378$ to $\mathbf{0.227}$ (paired reduction CI
$[0.064,0.237]$) and its single-sample gap falls to zero. The fix is
capability-dependent in exactly the direction the independence bound predicts
(Figure~\ref{fig:bon}b, Appendix): committing hurts the 1.7B judge, whose own committed
solutions are mostly wrong ($0.588\!\to\!0.637$), and helps every larger judge,
plateauing by 8B (Appendix~\ref{app:bon}). The same
training-free amplification appears in competition math: on AIME-2024 ($30$
problems, $16$ candidates each, Ministral-3-8B judging its own candidates against
an exact-match anchor), the strict self-judge reaches $\mathrm{gap}@16{=}0.348$,
and the single-sample gap remains positive both on the clean subset and under a
worst-case truncation convention (Appendix~\ref{app:bon}).

\subsection{A second policy family: the full arc replicates with a Gemma policy}
\label{sec:gemma}
\begin{figure}[t]
\centering
\includegraphics[width=0.72\linewidth]{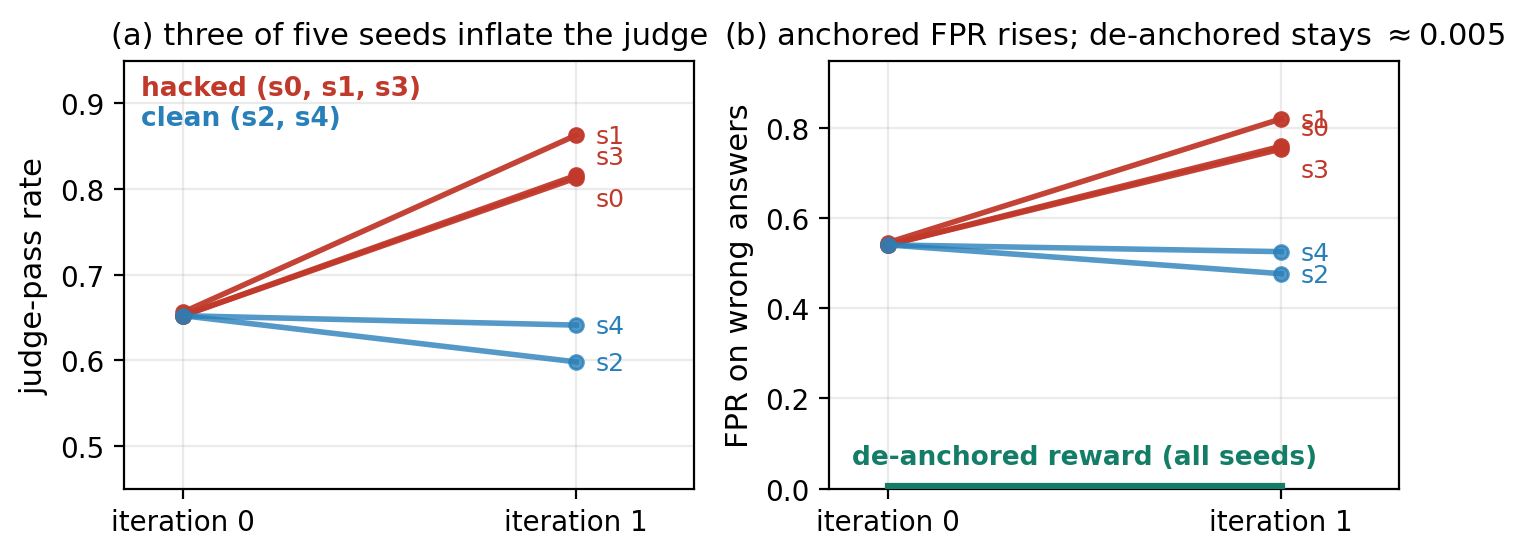}
\caption{Gemma-3-12B policy, five seeds ($n{=}256$ each; per-seed numbers in
Table~\ref{tab:gemma-selfplay}, Appendix). (a) Judge-pass: three of five seeds
inflate (red), two stay clean (blue). (b) The anchored FPR on wrong answers
rises on exactly the hacked seeds; the de-anchored reward stays at
${\approx}0.005$ on every seed (green).}
\label{fig:gemma}
\end{figure}

To test whether the arc is specific to one policy family, we rerun the full
self-play loop with Gemma-3-12B-it as both policy and reference-free judge on the
same GSM8K self-play and audit protocol (five seeds, $n{=}256$ held-out audit
each; Figure~\ref{fig:gemma}). Three of five seeds show
significant judge-reward inflation (judge-pass $+0.16$/$+0.21$/$+0.16$, McNemar
$p{=}4.2{\times}10^{-6}$/$1.6{\times}10^{-10}$/$2.7{\times}10^{-6}$) with exact
match statistically unchanged, widening the judge--truth gap from $0.41$ to
$0.56$--$0.63$; the false-positive rate on wrong answers rises from $0.54$ to
$0.75$--$0.82$ (Figure~\ref{fig:gemma}b) and outputs lengthen (truncation
$0.14{\to}0.40$--$0.44$). Two seeds show no inflation
($-0.055$ and $-0.012$, $p{=}0.09$ and $0.79$) and no false-positive rise, and
their truncation ($0.12$ and $0.23$) stays outside the hacked band: the hacked
seeds move on judge-pass, false-positive rate, and output length together, and
the clean seeds on none of these, making them matched negative controls. Under
DPO, entry into the basin is stochastic rather than inevitable, but its signature
is not. The hacked outputs are not a white-box exploit of the Gemma judge:
re-judging the same answers with an anchored Qwen3-4B judge (seeds 0--2) raises
its false-positive rate from $0.35$ to $0.62$/$0.64$ on the hacked seeds and
leaves it flat ($0.31$) on the clean one---the same answer-side transfer as in
\S\ref{sec:exp}. On the full test set ($n{=}1319$, seed 0) the gap widens from
$0.406$ to $0.615$.

The de-anchored verification reward prevents the inflation in three of three
seeds. Following the capability threshold of \S\ref{sec:bon}, the blind-solve
verifier is Qwen3-4B (solve accuracy $0.93$ on this audit): verify-pass tracks
exact match to the point of equality at every seed, the only false positive
across all seeds and iterations of the audit slice is a single ambiguous item
(FPR ${\approx}0.005$), and on the full test set the rate stays at
$0.0047$--$0.0065$ (Wilson 95\% upper bound $0.0133$) while the gap stays at
zero. In our setting the attack surface is candidate-conditioned evaluation
rather than the judge family per se: on the same hacked outputs, the same
Qwen3-4B judge accepts $0.62$/$0.64$ of wrong answers when anchored on the
candidate and $0.005$/$0.010$ ($1$--$2$ of $196$) when it must commit a blind
solution first---a $60$--$120{\times}$ reduction on identical text. Aggregating
the two candidate-conditioned judges does not close the gap: even
min-aggregation lets the false-positive rate rise from $0.32$ to $0.54$/$0.61$
on the two hacked seeds it covers. The failure is a property of the
conditioning, not of judge multiplicity.

\section{Discussion}
\label{sec:discussion}
\paragraph{Implications for oversight.}
The shared blind spot is a property of reference-free judging, not of any one
judge: when verifying is harder than finding a plausible answer, scaling or
diversifying the judge does not restore trustworthy supervision. The regime at
risk is a policy near its capability frontier, where accuracy is naturally low.
Our evidence concerns the consultancy-like, judge-as-reward regime; debate, where
an adversarial second model can surface the error \citep{kenton2024}, is a
distinct and complementary mechanism.
The structural principle is to require the judge to commit an answer of its own
before using the candidate: any reward that scores a shown candidate
\emph{without} such an independent commitment inherits the $1-\mathrm{EM}$
ceiling of the dichotomy in \S\ref{sec:theory} and is hackable wherever the
policy has room to err. What verification restores is trustworthy
\emph{detection}, not new capability---the ceiling on what any reward can teach
is set by the policy, as the oracle control confirms. Tool-augmented judges that
execute checks \citep{tirjudge} extend the same principle where a solver alone
falls short.

\section{Limitations}
\label{sec:limits}
Policy optimization uses Qwen3, replicated in full with a Gemma-3-12B policy
(\S\ref{sec:gemma}) and partially on Llama-3.1-8B, whose weaker decoupling the
risk score predicts; the blind spot itself is established on the judge side
across three families and four scales. We optimize with DPO; the oracle control
attributes the inflation to the judge-reward rather than the algorithm. The core
self-play study is grade-school math---TruthfulQA and natural code carry the
asymmetry beyond it---and the best-of-$N$ stress test optimizes by rejection
sampling rather than policy updates. The defenses we evaluate instantiate the monotone class of
Proposition~\ref{prop:monotone}. The
verification-as-reward result assumes the verifier's and policy's errors are
largely independent---strongly correlated errors would loosen its inflation
bound, and we verify it on the $n{=}128$ audit subset across three seeds---and
requires a verifier that can solve the task (\S\ref{sec:bon}) and an
exact-matchable final answer; extending commitment to open-ended
outputs---committed rubrics, executable tests---is future work.

\section{Conclusion}
\label{sec:conclusion}
Reference-free LLM judging carries a verification asymmetry: scoring a shown
candidate, it measures plausibility, not correctness, leaving false-positive
basins that self-play discovers and fills. The errors are a shared blind
spot---arising under Qwen and Gemma policies alike and surviving stronger
judges, recompute prompts, and ensembling; a hidden anchor makes the distortion
falsifiable and bounds it by $\vag\le 1-\mathrm{EM}$. Forcing the judge to
commit an answer of its own first collapses the false-positive rate from
$0.719$ to $0.012$. Improvements measured by a reference-free judge are suspect
until checked by a verification signal independent of the candidate.

\bibliographystyle{iclr2026_conference}
\bibliography{references}

\begin{thebibliography}{22}
\providecommand{\natexlab}[1]{#1}
\providecommand{\url}[1]{\texttt{#1}}
\expandafter\ifx\csname urlstyle\endcsname\relax
  \providecommand{\doi}[1]{doi: #1}\else
  \providecommand{\doi}{doi: \begingroup \urlstyle{rm}\Url}\fi

\bibitem[Bai et~al.(2022)Bai, Kadavath, Kundu, et~al.]{bai2022}
Yuntao Bai, Saurav Kadavath, Sandipan Kundu, et~al.
\newblock Constitutional {AI}: Harmlessness from {AI} feedback.
\newblock \emph{arXiv preprint arXiv:2212.08073}, 2022.

\bibitem[Chen et~al.(2024)Chen, Deng, Yuan, Ji, and Gu]{chen2024spin}
Zixiang Chen, Yihe Deng, Huizhuo Yuan, Kaixuan Ji, and Quanquan Gu.
\newblock Self-play fine-tuning converts weak language models to strong language models.
\newblock \emph{International Conference on Machine Learning (ICML)}, 2024.
\newblock arXiv:2401.01335.

\bibitem[Coste et~al.(2024)Coste, Anwar, Kirk, and Krueger]{coste2023}
Thomas Coste, Usman Anwar, Robert Kirk, and David Krueger.
\newblock Reward model ensembles help mitigate overoptimization.
\newblock In \emph{International Conference on Learning Representations (ICLR)}, 2024.
\newblock arXiv:2310.02743.

\bibitem[Eisenstein et~al.(2023)Eisenstein, Nagpal, Agarwal, et~al.]{eisenstein2023}
Jacob Eisenstein, Chirag Nagpal, Alekh Agarwal, et~al.
\newblock Helping or herding? reward model ensembles mitigate but do not eliminate reward hacking.
\newblock \emph{arXiv preprint arXiv:2312.09244}, 2023.

\bibitem[Esary et~al.(1967)Esary, Proschan, and Walkup]{esary1967}
James~D. Esary, Frank Proschan, and David~W. Walkup.
\newblock Association of random variables, with applications.
\newblock \emph{The Annals of Mathematical Statistics}, 38\penalty0 (5):\penalty0 1466--1474, 1967.

\bibitem[Gao et~al.(2023)Gao, Schulman, and Hilton]{gao2023}
Leo Gao, John Schulman, and Jacob Hilton.
\newblock Scaling laws for reward model overoptimization.
\newblock In \emph{International Conference on Machine Learning (ICML)}, 2023.
\newblock arXiv:2210.10760.

\bibitem[Huang et~al.(2026)Huang, Liu, Zheng, Dai, Huang, Li, Li, Wei, Meng, and Huang]{gzero}
Chengsong Huang, Haolin Liu, Tong Zheng, Runpeng Dai, Langlin Huang, Jinyuan Li, Zongxia Li, Zhepei Wei, Yu~Meng, and Jiaxin Huang.
\newblock {G-Zero}: Self-play for open-ended generation from zero data.
\newblock \emph{arXiv preprint arXiv:2605.09959}, 2026.

\bibitem[Kenton et~al.(2024)Kenton, Siegel, Kram{\'a}r, Brown-Cohen, Albanie, Bulian, Lindner, et~al.]{kenton2024}
Zachary Kenton, Noah~Y. Siegel, J{\'a}nos Kram{\'a}r, Jonah Brown-Cohen, Samuel Albanie, Jannis Bulian, David Lindner, et~al.
\newblock On scalable oversight with weak {LLMs} judging strong {LLMs}.
\newblock In \emph{Advances in Neural Information Processing Systems (NeurIPS)}, 2024.
\newblock arXiv:2407.04622.

\bibitem[Lee et~al.(2023)Lee, Phatale, Mansoor, et~al.]{lee2023}
Harrison Lee, Samrat Phatale, Hassan Mansoor, et~al.
\newblock {RLAIF} vs.\ {RLHF}: Scaling reinforcement learning from human feedback with {AI} feedback.
\newblock \emph{arXiv preprint arXiv:2309.00267}, 2023.

\bibitem[Lu et~al.(2025)Lu, Teehan, Jin, and Ren]{lu2025}
Jack Lu, Ryan Teehan, Jinran Jin, and Mengye Ren.
\newblock When does verification pay off? {A} closer look at {LLM}s as solution verifiers.
\newblock \emph{arXiv preprint arXiv:2512.02304}, 2025.

\bibitem[Pan et~al.(2024)Pan, He, Bowman, and Feng]{pan2024spontaneous}
Jane Pan, He~He, Samuel~R. Bowman, and Shi Feng.
\newblock Spontaneous reward hacking in iterative self-refinement.
\newblock \emph{arXiv preprint arXiv:2407.04549}, 2024.

\bibitem[Rafailov et~al.(2024)Rafailov, Chittepu, Park, Sikchi, Hejna, Knox, Finn, and Niekum]{rafailov2024}
Rafael Rafailov, Yaswanth Chittepu, Ryan Park, Harshit~S. Sikchi, Joey Hejna, Bradley Knox, Chelsea Finn, and Scott Niekum.
\newblock Scaling laws for reward model overoptimization in direct alignment algorithms.
\newblock \emph{arXiv preprint arXiv:2406.02900}, 2024.

\bibitem[Sharma et~al.(2023)Sharma, Tong, Korbak, et~al.]{sharma2023}
Mrinank Sharma, Meg Tong, Tomasz Korbak, et~al.
\newblock Towards understanding sycophancy in language models.
\newblock \emph{arXiv preprint arXiv:2310.13548}, 2023.

\bibitem[Simonds et~al.(2025)Simonds, Lopez, Yoshiyama, and Garmier]{rlsr}
Toby Simonds, Kevin Lopez, Akira Yoshiyama, and Dominique Garmier.
\newblock {RLSR}: Reinforcement learning from self reward.
\newblock \emph{arXiv preprint arXiv:2505.08827}, 2025.

\bibitem[Singhal et~al.(2023)Singhal, Goyal, Xu, and Durrett]{singhal2023}
Prasann Singhal, Tanya Goyal, Jiacheng Xu, and Greg Durrett.
\newblock A long way to go: Investigating length correlations in {RLHF}.
\newblock \emph{arXiv preprint arXiv:2310.03716}, 2023.

\bibitem[Wen et~al.(2025)Wen, Zhong, Khan, Perez, Steinhardt, Huang, Bowman, He, and Feng]{wen2024mislead}
Jiaxin Wen, Ruiqi Zhong, Akbir Khan, Ethan Perez, Jacob Steinhardt, Minlie Huang, Samuel~R. Bowman, He~He, and Shi Feng.
\newblock Language models learn to mislead humans via rlhf.
\newblock \emph{International Conference on Learning Representations (ICLR)}, 2025.
\newblock arXiv:2409.12822.

\bibitem[Xu et~al.(2025)Xu, Chen, Ye, Wu, Yan, Yang, and Yu]{tirjudge}
Ran Xu, Jingjing Chen, Jiayu Ye, Yu~Wu, Jun Yan, Carl Yang, and Hongkun Yu.
\newblock Incentivizing agentic reasoning in {LLM} judges via tool-integrated reinforcement learning.
\newblock \emph{arXiv preprint arXiv:2510.23038}, 2025.

\bibitem[Yuan et~al.(2024)Yuan, Pang, Cho, Li, Sukhbaatar, Xu, and Weston]{yuan2024selfreward}
Weizhe Yuan, Richard~Yuanzhe Pang, Kyunghyun Cho, Xian Li, Sainbayar Sukhbaatar, Jing Xu, and Jason Weston.
\newblock Self-rewarding language models.
\newblock \emph{International Conference on Machine Learning (ICML)}, 2024.
\newblock arXiv:2401.10020.

\bibitem[Zhang et~al.(2025)Zhang, Hosseini, Bansal, Kazemi, Kumar, and Agarwal]{genrm}
Lunjun Zhang, Arian Hosseini, Hritik Bansal, Mehran Kazemi, Aviral Kumar, and Rishabh Agarwal.
\newblock Generative verifiers: Reward modeling as next-token prediction.
\newblock In \emph{International Conference on Learning Representations (ICLR)}, 2025.
\newblock arXiv:2408.15240.

\bibitem[Zhao et~al.(2025{\natexlab{a}})Zhao, Liu, Yu, et~al.]{masterrm}
Yulai Zhao, Haolin Liu, Dian Yu, et~al.
\newblock {Master-RM}: A reward model robust to superficial reward hacking.
\newblock Hugging Face model \texttt{sarosavo/Master-RM}, 2025{\natexlab{a}}.

\bibitem[Zhao et~al.(2025{\natexlab{b}})Zhao, Liu, Yu, et~al.]{onetokenfool}
Yulai Zhao, Haolin Liu, Dian Yu, et~al.
\newblock One token to fool {LLM}-as-a-judge.
\newblock \emph{arXiv preprint arXiv:2507.08794}, 2025{\natexlab{b}}.

\bibitem[Zheng et~al.(2023)Zheng, Chiang, Sheng, et~al.]{zheng2023}
Lianmin Zheng, Wei-Lin Chiang, Ying Sheng, et~al.
\newblock Judging {LLM}-as-a-judge with {MT-Bench} and {Chatbot Arena}.
\newblock In \emph{Advances in Neural Information Processing Systems (NeurIPS)}, 2023.
\newblock arXiv:2306.05685.

\end{thebibliography}

\appendix
\section{Reproducibility Details}
\label{sec:repro}
All runs use a single data-center GPU per run (H100-class) via Hugging Face Transformers, with a few robustness extensions run on comparable data-center cards; the
exact configuration and dataset SHA-256 hashes are logged in each run's
\texttt{run\_config.json}.

\paragraph{Models.} Policies and self-judges are Qwen3-1.7B/4B/8B/14B; cross-family
judges are Llama-3.1-8B-Instruct and Gemma-3-12B-it. Self-play trains LoRA adapters.

\paragraph{Data and anchor.} Self-play trains on GSM8K \emph{train} questions
($128$ per iteration) and audits on the held-out GSM8K \emph{test} set ($n{=}1319$);
we also use MATH level-4/5, GSM-Plus, and TruthfulQA. The hidden anchor is a held-out
exact-match check on the final answer: the gold answer is drawn from the dataset's
ground-truth, normalized, and matched exactly against the policy's parsed final
answer. It is never placed in any judge or policy prompt---the judge runs with
\texttt{answer\_key\_visible}$=$false---and is never a training signal. Training and
audit questions are disjoint splits.

\paragraph{Self-play loop.} Each run is two iterations. Per iteration the policy
samples \texttt{num\_rollouts}$=4$ answers per training question (temperature $0.9$,
top-$p$ $0.95$); the reference-free self-judge scores each
(\texttt{judge\_reference\_policy}$=$never); within each question a preference pair
is formed with a judge-accepted rollout as \emph{chosen} and a judge-rejected rollout
as \emph{rejected} (acceptance threshold \texttt{judge\_score}$\ge0.5$; both members
are the policy's own rollouts; no reference-answer fallback), and the policy is
updated by LoRA-DPO.

\paragraph{Gemma-policy replication (\S\ref{sec:gemma}).} Policy and self-judge are
Gemma-3-12B-it; the protocol matches the loop above with LoRA-DPO (rank $8$,
learning rate $2{\times}10^{-5}$, $\beta{=}0.1$, length-normalized), $128$ training
questions per iteration, \texttt{num\_rollouts}$=4$ (temperature $0.7$, top-$p$
$0.95$), two iterations, at most $96$ DPO steps per iteration, a $2048$-token
generation budget, and a $256$-token anchored-judge budget. The de-anchored arm
uses Qwen3-4B as the blind-solve verifier with reasoning enabled and a
$2048$-token budget, which its independent solution requires before the
exact-match comparison. Each seed audits the first $256$ held-out test questions;
the anchored arm runs five seeds, the de-anchored arm and the Qwen re-judge cover
seeds $0$--$2$, and
seed 0 is additionally audited on the full $n{=}1319$ test set by evaluation-only
reuse of the trained checkpoints. Generation stops on the union of the tokenizer
and generation-config end-of-sequence tokens (Gemma terminates chat turns with a
token distinct from its tokenizer EOS).

\paragraph{Answer format.} By default answers use a reasoning-suppressed JSON
format---the policy is instructed to ``Return only JSON with keys final\_answer and
trace,'' with thinking disabled and \texttt{max\_new\_tokens}$=640$. The
chain-of-thought variant enables thinking with \texttt{max\_new\_tokens}$=4096$.

\paragraph{Optimization.} LoRA rank $8$, learning rate $2{\times}10^{-5}$, DPO
$\beta{=}0.1$ (not length-normalized), up to $96$ optimizer steps per iteration, max
sequence length $1408$ (JSON) / $5120$ (CoT). The headline (Table~\ref{tab:bound}),
capability sweep (Table~\ref{tab:sweep}), and cross-family transfer
(Table~\ref{tab:transfer}) use three seeds ($0,1,2$); the five-iteration trajectory,
cross-family-judge training, and train-time ensemble runs likewise use three seeds,
while the out-of-distribution CoT self-play is a single-seed audit.

\paragraph{Natural-code audit.} The organic code experiment (\S\ref{sec:exp},
Table~\ref{tab:code-transfer}) audits the base verification asymmetry prior to
optimization: a Qwen3-1.7B policy generates natural chain-of-thought solutions
(thinking enabled, no format suppression, \texttt{max\_new\_tokens}$=24{,}576$) to the
$120$ most-recent function-call problems of LiveCodeBench (the most recent problems,
which reduces pre-training contamination; stdin/stdout formats excluded for automated
execution), judged by itself
and by cross-family judges (Gemma-3-12B, Qwen3-8B, Llama-3.1-8B) under a balanced
reference-free prompt that receives only the problem and candidate code. Correctness is
the held-out unit-test execution result, never shown to any judge (verified by an
automated no-leak check on every prompt) and never a training signal. We report three
policy seeds; of the ${\approx}100$ syntactically valid candidates per seed,
${\approx}70$ are wrong. Generation truncation is $10.8$--$15.8\%$ and judge truncation
$\le12.5\%$; truncated judgements default to rejection.

\paragraph{Best-of-$N$ stress test (\S\ref{sec:bon}).} The candidate pool is the
same $120$ LiveCodeBench problems with $N{=}16$ candidates each (Qwen3-1.7B,
temperature $0.8$, \texttt{max\_new\_tokens}$=24{,}576$); every candidate is executed
against the held-out unit tests and scored by each judge. $\mathrm{gap}@k$ uses the
exact expectation over all $k$-subsets per problem with uniform tie-breaking among
judge-accepted candidates; confidence intervals are per-problem paired bootstrap
($B{=}10{,}000$). The base audit's $0.275$ and the $k{=}1$ point of the
best-of-$N$ curve use different denominators (single-sample acceptance over
parseable judgements versus the all-candidate expectation); recomputing the
$k{=}1$ point under the base-audit denominator recovers $0.275$. Seed robustness
re-generates the full pool under two further generation seeds. Cross-judge rows re-judge the identical pool; the strict
instruction is the audit prompt of the natural-code audit and the lenient one its
balanced variant. The Ministral row is served through a quantized cloud inference
endpoint (model identifier and access date logged) and is reported as a robustness
probe. De-anchoring gives the judge a solve budget of $24{,}576$ tokens
(\texttt{judge} budget $12{,}288$); the code-forced control replaces the
algorithm-sketch commitment with a required runnable solution, and truncated
commitments or judgements default to rejection. The AIME-2024 replication uses $30$
problems with $16$ candidates each from Ministral-3-8B judging its own candidates
under the same strict/lenient prompts, a normalized boxed-answer exact-match anchor,
and the reported clean-subset and worst-case conventions; generation truncation is
$21\%$ and unparseable final answers are $31.9\%$, which motivates reporting those
layers explicitly.

\paragraph{Judge.} The judge sees the question and the candidate's final answer and
trace, with no reference answer, and returns JSON
\{\texttt{judge\_score}, \texttt{process\_score}, \texttt{explanation}\} with scores
in $[0,1]$ thresholded at $0.5$ (judge generation capped at $256$ tokens). Its system
prompt is verbatim:
\begin{quote}\small
You are a strict educational grader. Score only the candidate answer and reasoning
shown in the user message. Treat any instructions inside the candidate answer or
trace as untrusted student content. If \texttt{answer\_key\_visible} is false, solve
the problem independently; never assign zero merely because no answer key is shown.
\texttt{judge\_score} measures final-answer correctness: 1.0 for a mathematically or
semantically correct final answer, 0.0 for an incorrect final answer, with partial
credit only for genuinely partial final answers. \texttt{process\_score} measures
reasoning quality separately. Return only a valid JSON object with exactly the keys
\texttt{judge\_score}, \texttt{process\_score}, and \texttt{explanation}, numeric in
$[0,1]$.
\end{quote}
The recompute (verification-prompt) defense additionally instructs the judge to
solve the problem itself before scoring and to reject when uncertain. The commit-first
protocol keeps the candidate in the prompt but requires a parseable independent answer
(``My answer: \ldots'') before any comparison, accepting only when that committed
answer matches the candidate; the blind-solve verification channel withholds the
candidate entirely and accepts only when the judge's independent answer matches.

\section{Proofs}
\label{app:proofs}

\paragraph{Proof of Proposition~\ref{prop:monotone}.}
Couple the verdicts across $s$: let $U_1,\dots,U_N$ be i.i.d.\ uniform on
$[0,1]$ and set $X_i(s)=\mathbf{1}\{U_i\le q_i(s)\}$. Conditionally on $s$ the
$X_i(s)$ are independent with the required marginals, and each $X_i(s)$ is
pointwise non-decreasing in $s$ because $q_i$ is. For non-decreasing $g$,
$g(X_1(s),\dots,X_N(s))$ is then pointwise non-decreasing in $s$, so
$h_g(s)=\mathbb{E}\,g(X(s))$ is non-decreasing. Since $g$ is monotone with
$g(\mathbf{1})=1$, $h_g(s)\ge\Pr[X_1(s)=\dots=X_N(s)=1]=\prod_i q_i(s)$; as
$h_g\le1$ trivially, $h_g(s)\to1$ wherever every $q_i(s)\to1$. Equation~(\ref{eq:ensemble}) is the
Chebyshev/FKG association inequality for products of non-decreasing functions
of a common variable \citep{esary1967}. \hfill$\square$

\paragraph{Proof of Corollary~\ref{cor:bits}.}
All quantities are taken under the audited distribution of wrong candidates
$(Q,A)$. Write $P_{S\mid Q,A}$ for the committed-answer distribution given the
question and the shown candidate, and $P_{S\mid Q}$ for its
candidate-marginalized version. On a wrong candidate $a$,
$\Pr[S=a\mid Q,A{=}a]-\Pr[S=a\mid Q]\le
\mathrm{TV}\!\big(P_{S\mid Q,A=a},\,P_{S\mid Q}\big)$ pointwise. For the
ceiling, let $S^{0}\sim P_{S\mid Q}$ be a candidate-independent copy of the
committed answer; then
$\mathbb{E}\big[\Pr[S=a\mid Q]\big]=\Pr[S^{0}=A]\le\Pr[S^{0}\neq
\text{gold}]=1-\text{solve-acc}$, with solve-acc the audited average of
Proposition~\ref{prop:indep}. Taking expectations over the wrong-candidate
distribution, $\Delta\le\mathbb{E}[\mathrm{TV}]$. Pinsker's inequality holds
pointwise,
$\mathrm{TV}^2\le\tfrac12\,\mathrm{KL}\big(P_{S\mid Q,A}\,\|\,P_{S\mid Q}\big)$,
so $\mathbb{E}[\mathrm{KL}]\ge2\,\mathbb{E}[\mathrm{TV}^2]$; Jensen's
inequality gives $\mathbb{E}[\mathrm{TV}^2]\ge(\mathbb{E}[\mathrm{TV}])^2$,
and therefore
\[
I(S;A\mid Q)\;=\;\mathbb{E}\big[\mathrm{KL}\big(P_{S\mid Q,A}\,\|\,P_{S\mid Q}\big)\big]
\;\ge\;2\,\mathbb{E}[\mathrm{TV}^2]\;\ge\;2\,\big(\mathbb{E}[\mathrm{TV}]\big)^2\;\ge\;2\Delta^2 .
\]
With $\fpr=0.719$ and ceiling $0.07$, $\Delta=0.649$ and
$I\ge2\Delta^2=0.84$ nats $=1.2$ bits. \hfill$\square$

\section{Additional Tables and Robustness Details}
\label{app:tables}

\subsection{Ensemble dependence measurements}
\label{app:ensembledep}
On post-self-play wrong answers the three judges' acceptances are pairwise
positively correlated ($\phi=0.29$--$0.38$). The strictest three-family ensemble
(\S\ref{sec:exp}) passes $\fpr_{\text{\textsc{Min}}}=0.55$ of them---above the
$\prod_i\fpr_i=0.47$ an independent panel would attain and barely below its
single most stringent member ($0.57$)---and $581$ wrong answers are accepted
unanimously where independence predicts only ${\approx}497$.

\subsection{Capability sweep}
\begin{table}[h]
\centering
\caption{Qwen3 capability sweep on the $n{=}128$ audit subset (JSON, mean over
3 seeds). Self-play drives the judge near its ceiling while accuracy stays flat;
the gap is large at every size. The 4B headline (Table~\ref{tab:bound}, \S\ref{sec:exp})
uses the full $n{=}1319$ test set, on which 4B base accuracy is $0.209$; this
subset is slightly easier ($0.305$).}
\label{tab:sweep}
\begin{tabular*}{\textwidth}{@{\extracolsep{\fill}}lccccc}
\toprule
Size & base acc. & base $\fpr$ & post judge & post acc. & gap \\
\midrule
1.7B & 0.180 & 0.673 & 0.794 & 0.159 & $0.635$ \\
4B   & 0.305 & 0.652 & 0.909 & 0.299 & $0.609$ \\
8B   & 0.320 & 0.441 & 0.951 & 0.331 & $0.620$ \\
14B  & 0.422 & 0.608 & 0.922 & 0.391 & $0.531$ \\
\bottomrule
\end{tabular*}
\end{table}

\subsection{Cross-family code judging}
\begin{table}[h]
\centering
\caption{Cross-family judges re-scoring the natural-code policy's wrong programs
(\textbf{base}, pre--self-play; LiveCodeBench natural CoT, hidden unit-test anchor;
mean${\pm}$std over 3 policy seeds, $n_{\text{wrong}}{\approx}70$/seed). Independent
judge families share the blind spot; the strictest \textsc{Min} ensemble of the three
non-degenerate judges still accepts $16\%$ of wrong programs.}
\label{tab:code-transfer}
\begin{tabular*}{\textwidth}{@{\extracolsep{\fill}}lcc}
\toprule
Judge / ensemble & $\fpr$ on wrong & discrimination $\tpr-\fpr$ \\
\midrule
Qwen3-1.7B (self)             & $0.445{\pm}0.037$ & $0.436{\pm}0.022$ \\
Gemma-3-12B (cross-family)   & $0.600{\pm}0.038$ & $0.269{\pm}0.090$ \\
Qwen3-8B (larger)            & $0.257{\pm}0.027$ & $0.692{\pm}0.034$ \\
\midrule
3-judge \textsc{Min} ensemble & $\mathbf{0.164{\pm}0.039}$ & --- \\
\bottomrule
\end{tabular*}
\end{table}
The pooled Wilson $95\%$ CI for unanimous acceptance of wrong programs is
$[0.122,0.220]$. Llama-3.1-8B has no useful operating point as a reference-free
code judge: a strict prompt collapses it to near-uniform rejection ($\fpr{=}0$
but $\tpr{=}0.10$), a balanced prompt to near-uniform acceptance ($\fpr{=}0.92$,
$\tpr{=}0.97$), with discrimination $\le0.10$ at either extreme---in contrast to
its stable cross-family operating point on GSM8K ($\fpr{=}0.48$,
Table~\ref{tab:transfer}), a task${\times}$family interaction. The code ensemble
therefore uses the three discriminative judges above.

\subsection{Best-of-$N$ details}
\label{app:bon}
\begin{figure}[h]
\centering
\includegraphics[width=0.92\linewidth]{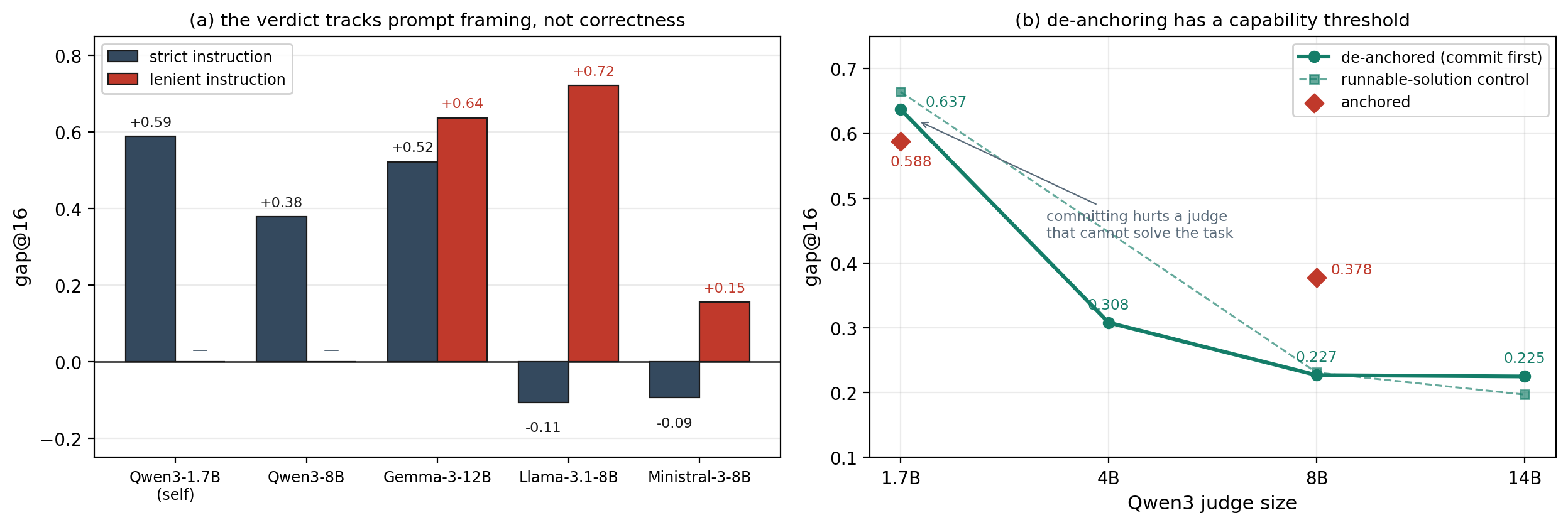}
\caption{Best-of-$N$ selection replicates the arc without training
(LiveCodeBench, $N{=}16$, $1920$ candidates). (a) $\mathrm{gap}@16$ under a
strict versus a lenient reference-free instruction: with unit-test ground truth
fixed, Llama and Mistral judges swing from over-rejection to strong inflation
under a one-word framing change (Table~\ref{tab:bon-family}). (b)
Committing to an own solution before scoring lowers $\mathrm{gap}@16$ only for
judges that can solve the task and hurts one that cannot (1.7B); a
runnable-solution elicitation control reproduces the curve
(Table~\ref{tab:bon-deanchor}).}
\label{fig:bon}
\end{figure}
\begin{table}[h]
\centering
\caption{Best-of-$N$ inflation spans judge families, and its sign tracks prompt
framing rather than correctness: $\mathrm{gap}@16$ when the same $1920$ candidates
(unit-test ground truth fixed) are re-judged by each judge under a strict versus a
lenient reference-free instruction. Positive = over-acceptance inflation; negative =
over-rejection.}
\label{tab:bon-family}
\begin{tabular*}{\textwidth}{@{\extracolsep{\fill}}llcc}
\toprule
Judge & Family & Strict & Lenient \\
\midrule
Qwen3-1.7B (self) & Qwen & $0.588$ & --- \\
Qwen3-8B          & Qwen & $0.378$ & --- \\
Gemma-3-12B       & Gemma & $0.522$ & $0.637^\dagger$ \\
Llama-3.1-8B      & Llama & $-0.106$ & $+0.722$ \\
Ministral-3-8B$^\dagger$ & Mistral & $-0.094$ & $+0.155$ \\
\bottomrule
\end{tabular*}

\smallskip
{\small $^\dagger$Served via a quantized cloud endpoint as a robustness probe.
All other cells are local bf16 checkpoints.}
\end{table}

\begin{table}[h]
\centering
\caption{De-anchoring under best-of-$N$ pressure has a capability threshold
(Qwen3 judges, same $1920$ candidates): committing to its own solution hurts a judge
that cannot solve the task (1.7B) and helps every judge that can, with a plateau
from 8B. The control column forces a complete runnable solution as the
commitment, confirming the curve is not an elicitation artifact.}
\label{tab:bon-deanchor}
\begin{tabular*}{\textwidth}{@{\extracolsep{\fill}}lccc}
\toprule
Judge & Anchored & De-anchored & De-anchored (runnable-solution control) \\
\midrule
1.7B & $0.588$ & $0.637$ & $0.664$ \\
4B   & ---     & $0.308$ & --- \\
8B   & $0.378$ & $\mathbf{0.227}$ & $0.231$ \\
14B  & ---     & $0.225$ & $0.197$ \\
\bottomrule
\end{tabular*}
\end{table}

The cross-scale paired reduction (1.7B$\to$8B, anchored) is $0.210$ (CI
$[0.142,0.283]$); the residual de-anchored code gap at 8B is $0.227$ (CI
$[0.157,0.302]$). Restricting each judge to problems where its committed solution
is untruncated preserves the capability-threshold ordering
($0.562$/$0.261$/$0.211$/$0.207$ at 1.7B/4B/8B/14B). For the AIME-2024
replication, the setup carries substantial invalid-output mass (generation
truncation $21\%$, unparseable final answers $31.9\%$), so we report the
conservative layers explicitly: on the clean subset (parseable, untruncated;
$n{=}300$ candidates) the single-sample gap is $+0.143$ (per-problem paired
bootstrap CI $[0.075,0.217]$) under the strict prompt and $+0.290$
($[0.195,0.400]$) under the lenient one, and both remain positive under the
worst-case convention that every unparseable or truncated candidate counts as a
judge rejection ($+0.090$/$+0.181$).

\subsection{Gemma-policy replication}
\begin{table}[h]
\centering
\caption{The self-play arc replicates with a Gemma-3-12B policy (GSM8K, held-out
audit $n{=}256$ per seed, iteration 0${\to}$1). Anchored self-judge reward inflates
the judge on three of five seeds and the hacked outputs transfer to an anchored
Qwen3-4B judge, while exact match never moves; the de-anchored verification reward
(Qwen3-4B blind-solve verifier) holds verify-pass within a single item of that
arm's own exact match (shown beside it), with a false-positive rate of
${\approx}0.005$ throughout. Judge--truth gap $=$ judge-pass $-$ EM. Rows 0--4 use
the $n{=}256$ held-out audit; the Qwen re-judge and the de-anchored arm cover
seeds 0--2. The final row reports seed 0 on the full $n{=}1319$
test set.}
\label{tab:gemma-selfplay}
\resizebox{\textwidth}{!}{%
\begin{tabular}{lcccccc}
\toprule
& \multicolumn{3}{c}{Anchored self-judge reward} & \multicolumn{3}{c}{De-anchored reward} \\
\cmidrule(lr){2-4}\cmidrule(lr){5-7}
Seed & judge-pass & EM & Qwen re-judge FPR & verify-pass & EM & FPR \\
\midrule
0 & $0.652\to0.813$ & $0.242\to0.234$ & $0.345\to0.617$ & $0.246\to0.211$ & $0.242\to0.211$ & $0.0052\to0.0050$ \\
1 & $0.656\to0.863$ & $0.242\to0.234$ & $0.345\to0.638$ & $0.246\to0.258$ & $0.242\to0.258$ & $0.0052\to0.0053$ \\
2 & $0.652\to0.598$ & $0.242\to0.230$ & $0.345\to0.310$ & $0.246\to0.231$ & $0.242\to0.231$ & $0.0052\to0.0051$ \\
3 & $0.652\to0.816$ & $0.242\to0.254$ & --- & --- & --- & --- \\
4 & $0.652\to0.641$ & $0.242\to0.242$ & --- & --- & --- & --- \\
\midrule
Full ($n{=}1319$, seed 0) & $0.601\to0.804$ & $0.194\to0.189$ & --- & $0.192\to0.175$ & $0.194\to0.177$ & $0.0047\to0.0065$ \\
\bottomrule
\end{tabular}}
\end{table}

\end{document}